\documentclass[letterpaper]{article}

\usepackage{natbib,alifeconf}  

\usepackage{url}

\title{Evolutionary Innovation Viewed as Novel Physical Phenomena \\ and Hierarchical
  Systems Building}
\author{Tim Taylor$^{1,2}$ \\
  \mbox{}\\
  $^1$Independent Researcher \\
  $^2$Department of Data Science and AI, Monash University \\
tim@tim-taylor.com} 

\begin{document}

\newcommand{\oto}{\mbox{1-to-1}}

\maketitle


\section{Introduction}

In previous work I proposed a framework for thinking about open-ended
evolution \citep{taylor2019evolutionary}. The framework characterised
the basic processes required for Darwinian evolution as: (1) the
generation of a phenotype from a genetic description; (2) the
evaluation of that phenotype; and (3) the reproduction with variation
of successful genotype-phenotypes. My treatment emphasized the
potential influence of the biotic and abiotic environment, and of the
laws of physics/chemistry, on each of these processes. I demonstrated
the conditions under which these processes can allow for ongoing
exploration of a space of possible phenotypes (which I labelled
\emph{exploratory open-endedness}).
However, these processes by themselves cannot
\emph{expand the space of possible phenotypes} and therefore cannot account
for the more interesting and unexpected kinds of evolutionary
innovation (such as those I labelled \emph{expansive} and \emph{transformational}
open-endedness\footnote{See \citep{taylor2019evolutionary} for a full
  description of these three kinds of innovations and the differences
  between them.}). 

In the previous work I looked at ways in which \emph{expansive} and
\emph{transformational} innovations could arise. I proposed
transdomain bridges and non-additive compositional systems as two
mechanisms by which these kinds of innovations could arise. In the
current paper I wish to generalise and expand upon these two
concepts. I do this by adopting the Parameter Space--Organisation
Space--Action Space (POA) perspective, as suggested at the end
of \citep{taylor2019evolutionary}, and proposing that all evolutionary
innovations can be viewed as either capturing some novel physical
phenomena that had previously been unused, or as the creation of new
persistent systems within the environment.

\section{Parameter Space, Organisation Space and Action Space}
In any system (real or virtual) governed by global laws of physics,
dynamics come about through the action of these laws upon the matter (objects)
within the system. If the behaviour of the system is determined solely by the
initial conditions of the system and by the global laws of physics
that act upon it, there appears to be little room for \emph{agency},
i.e.\ for organisms within the system that appear to follow their own
rules of behaviour. 
However, the behaviour in any given subregion of space at any given
time is
affected by local constraints.\footnote{\cite{montevil2015biological}
  define constraints as ``contingent causes, exerted by specific
  structures or dynamics, which reduce the degrees of freedom of the
  system in which they act.'' Some authors in this area tend to use
  the term \emph{boundary condition} rather than constraint, e.g.\
  \citep{kauffman2020answering},
  and others use both terms, e.g.\ \citep{pattee2015cell}. Here I use
  the term \emph{constraint} in a general way that also encompasses the
  concept of boundary conditions.} 
At the most fundamental level, the way in which living organisms ``break
away'' from these global laws of physics and seemingly follow their
own rules of behaviour is by creating local constraints on
dynamics from information stored in their genomes
\citep{pattee1972laws}.\footnote{At a higher level, some living 
systems also store information in cognitive memories or in cultural or
environmental memories too. In this paper I restrict my attention to
information stored in the genome.} This genetic information accumulates over
evolutionary time through the action of natural selection upon
organisms.

\begin{figure}[t]
\begin{center}
\includegraphics[width=\linewidth]{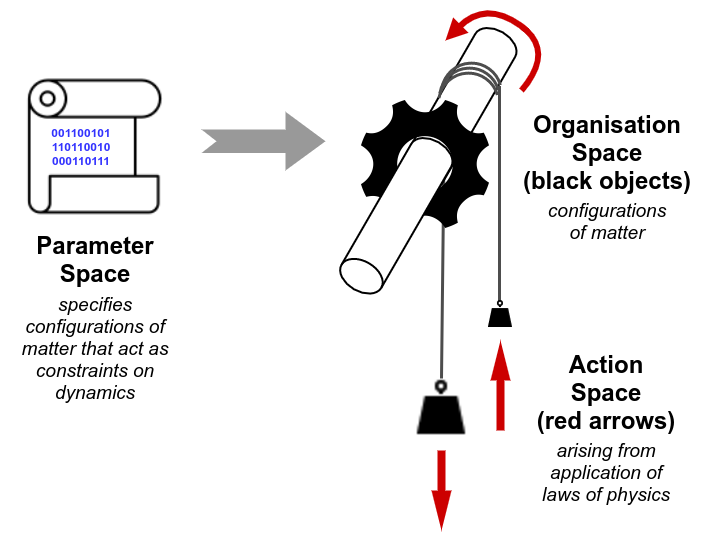}
\caption{Parameter Space, Organisation Space and Action Space. In this
mechanical example, the information in parameter space specifies a material
configuration of weights, rope, an axle and a cogwheel. This
configuration constrains the system's dynamics arising from the
application of the laws of physics to produce the rotation of the
cogwheel.} 
\label{fig-POA}
\end{center}
\end{figure}

Understanding how novel organismic behaviours might arise in an evolutionary
system first requires an appropriate conception of what a behaviour
is. In the context just described, we can say that an organism's
behaviour in a (real or virtual) physical system comprises three
core aspects:
(1) \emph{information} stored in a persistent memory that can
generate\footnote{At the most basic level, sections of the memory act
  directly as constraints upon the local dynamics. In more
  highly evolved systems such as modern biological organisms, there
  might be a complex hierarchical system acting as an interpreter of
  the information stored in the memory. These more complex cases
  are discussed later in this paper.} 
(2) specific \emph{configurations of matter} (constraints) that (3) are acted
upon by the global \emph{laws of physics} to produce specific actions.
Accordingly, behaviour can be viewed as comprising components in three
distinct but interacting spaces, which I call \emph{Parameter Space},
\emph{Organisation Space}, and \emph{Action Space}, respectively (see
Figure~\ref{fig-POA}). I refer to this as the \emph{POA} perspective.

These concepts can be applied not just to real and virtual physical
systems, but (at least partially) to more abstract artificial worlds
too. For example, a program in \emph{Tierra} \citep{ray1991approach} comprises a
linear list of symbols---it is an object in Organisation Space. The program
produces a behaviour in Action Space when acted upon by the CPU. In
this simple 1-dimensional world, there is a \oto\ mapping between
Parameter Space and Organisation Space, although modifications of
Tierra have been developed in which that is not the case, e.g.\
\citep{baugh2013evolution}. 

If we adopt this perspective, where the information contained in an
organism's genome is ``merely'' constraining and sculpting dynamics
caused by the application of global laws of physics, what
opportunities exist for the emergence of evolutionary innovations?
This question will be addressed in the following sections.

\section{Evolutionary Innovations via Novel Physical Phenomena}
In the kind of system described above, an evolutionary innovation can
arise if an organism generates constraints that trigger a physical
phenomenon that was not previously utilised.

A classic example is a wing (an object in Organisation Space, specified
by the organism's genetic information in Parameter Space). When acted
upon by the physical laws of aerodynamics, the wing produces uplift
and enables flight (in Action Space).

Another classic example is a photosensitive pigment such a rhodopsin
(an object in Organisation Space, specified by the organism's genetic
information in Parameter Space) that, following the laws of molecular
physics, changes its shape when it absorbs a photon (in Action Space).
In so doing, it can act as the basis for photosensitivity and vision. 

These kinds of evolutionary innovations, based upon capturing new
physical phenomena, are generally absent from computational
evolutionary systems.\footnote{The closest we could get to finding
  a novel physical phenomenon in a Tierra-like world would be if we
  provided an instruction in the instruction set that was unused in
  the ancestor organism (for example, an instruction to read the
  system time), but which appeared in a descendant by a fortuitous
  mutation---assuming the mutation operator allowed for this.} This is
because objects in computational systems 
usually only have a single domain of interaction (i.e.\ only one class
of laws of physics apply to them). Objects in the
real physical world, on the other hand, possess multiple properties
across various domains of interaction (mechanical, electrical,
chemical, etc.) and can thereby act as \emph{transdomain bridges}
\citep{taylor2019evolutionary}---that is the reason why they can
represent exaptations for innovations in different domains.

It is noteworthy that we \emph{do} see these kinds of innovations in
artificial evolutionary systems that are embedded in the real world
rather than being purely virtual systems. Examples include Pask's
evolved ``ear'' \citep{cariani1993evolve} and \cite{bird2002evolved}'s
evolved radio receiver. In these cases the system's components, being
physical objects, \emph{do} have properties in multiple domains of
interaction and therefore \emph{can} act as transdomain bridges.

\section{Evolutionary Innovations via (Hierarchical) Systems Building}
Even without the discovery of new physical phenomena, an evolutionary
system can produce innovations by combining existing components in
novel ways. The threefold view of innovations adopted in
\citep{taylor2019evolutionary} was based upon
\cite{banzhaf2016defining}'s proposal. That proposal took a systems
view of evolutionary novelties. However, even though the authors
defined an innovation as ``a change that adds a new \emph{type} or
\emph{relationship}'' \citep[p.142]{banzhaf2016defining} (emphasis
added), their discussion tended to focus specifically upon the
introduction of new \emph{entities} into a system. Here I would like
to build upon this systems view of evolutionary novelty, and to
emphasize the role of \emph{new relationships} in generating novelties
too. The potential benefit of doing this is nicely described by
philosopher William Bechtel as follows:

\begin{quotation}
  What is possible when components are put together in creative ways
  is often obscured when one focuses just on the components
  themselves\ldots\ One can appreciate this point by turning one's
  focus from science to engineering. Engineers do not build new
  devices by creating matter with distinctive properties \emph{ab
  initio}. Rather, they start with things that already exist and put
  them together in novel ways. What can be accomplished when the parts
  are put together is typically far from
  obvious. \citep[p.277]{bechtel2007biological}  
\end{quotation}

\cite{lenton2021survival} have recently argued that natural selection
can occur in systems above reproducing individuals (e.g.\ ecosystems)
based solely upon differential persistence. Persistence in these
higher-level systems is largely driven by ``self-perpetuating feedback
cycles involving biotic as well as abiotic components''
\citep[p.333]{lenton2021survival}. This view offers the possibility of
a consistent understanding of all levels of biological life, from the
cellular level to the levels of cultures, ecosystems and
biogeochemical cycles, in terms of the evolution of persistent
systems, with higher-level complex systems being hierarchically
composed of a series of lower-level systems.

In addition to entities (components) and feedback cycles, various
proposals exist for a common vocabulary for describing the general
structure and operation of biological, technical and cultural
systems. For instance,
\cite{rosnay1975macroscope} describes the principal \emph{structural}
characteristics of a system in terms of its \emph{limits} (boundaries),
\emph{elements}, \emph{reservoirs} and \emph{communication networks}
(the latter two applying to energy, material and information within the
system). Furthermore, the principal \emph{functional} characteristics are
described in terms of \emph{flows} (of energy, material and information),
\emph{valves and regulators}, \emph{delays}, and \emph{feedback loops}.

De Rosnay (\citeyear{rosnay1975macroscope})'s attempt to develop a general
vocabulary and visual representation for describing the structure and
function of systems is perhaps more all-encompassing than approaches
typically found in cybernetics, e.g.\ \citep{wiener1948cybernetics},
or in general system theory, e.g.\ \citep{bertalanffy1968general}.
The kind of approach set out by de Rosnay has more recently been
developed into separate notations for diagramming a system's stocks
and flows and its causal loops, e.g.\ \citep{sterman2000business}.
Attempts to develop a general language for describing system structure
and function can also be found in other disciplines, such as
engineering, e.g.\ \citep{pahl1988engineering}, and artificial life,
e.g.\ \citep{grand2000creation}.
While none of these proposals feels like the last word on the subject,
they nevertheless hold out the prospect of a universal language for
describing systems, comprising a relatively small set of common
structural and functional elements.

The lines of development described above suggest that it might be
possible to adopt an all-encompassing view of evolutionary
innovations as \emph{the creation and modification of information-driven
persistent systems in the environment}. Such a view could be applied
to innovations at all levels, from within individual organisms to
higher-level systems involving multiple biotic and abiotic elements
and cycles. Specific innovations could be classified in terms of a
universal set of structural and functional elements, and in terms of
whether the innovation represents:
\begin{enumerate}
  \item A change to an existing structural or functional element
    within a system, 
  \item An addition of a new element (or deletion of an existing
    element), or 
  \item The creation of a new hierarchical level of system built upon
    existing lower-level systems. 
  \end{enumerate}
To give a few examples: a Type 1 innovation might be a change in a regulatory
connection in a genetic regulatory network resulting in the production
of an additional copy of a body part; a Type 2 innovation might be an
addition of a new species into an ecosystem; and a Type 3 innovation
might be a major transition in individuality from single-celled
to multicellular organisms.

While the importance of systems and feedback has long been recognised
at various levels of life, from genetic regulatory networks to
ecosystems,
an all-encompassing
systems view of the kind described above could highlight a much
broader range of commonalities among innovations in different kinds of
systems. 


\section{The Generation of Phenotypes}
Earlier, when discussing how information stored in a
persistent memory specifies configurations of matter that act as
constraints upon a system's dynamics, I remarked that this
specification might be a simple \oto\ mapping or it may be more
complex. As mentioned, a \oto\ mapping exists in ALife systems of
linear code such as Tierra. One step up in complexity would be to have
the laws of physics act upon a linear information string to generate a
specific constraint, as happens in the self-organised folding of RNA
molecules to form ribozymes. But of course the process of generating
phenotypes from genotypes in modern organisms is much more complicated
than that, and can be viewed as a complex hierarchical system of
constraints and dynamics initiated by the genetic information that
ultimately cycles back to control the transcription and translation of
that information.


The generation of phenotype from genotype therefore involves both the
POA perspective (information specifying constraints upon dynamics) and
the Systems perspective (in terms of how the genetic information is
ultimately interpreted as complex constraints and dynamics). In
modern organisms it is a complex system 
entailing the processing of information, matter and energy based upon
low-level self-assembly processes but using those basic levels to
build further hierarchical layers of more complex structures and
dynamics.
This view of complex phenotypes being generated by
hierarchically structured interpretation processes that at the base
level utilise dynamics governed by self-organisation and other laws of
physics/chemistry but at higher levels utilise non-elementary tasks
constructed from the lower level systems, is echoed in recent work on
\emph{constructor theory} \citep{marletto2015constructor} and in
\cite{sloman2017construction}'s concepts of \emph{fundamental} versus
\emph{derived} construction kits.

As the generation of phenotype from
genotype is handled by increasingly more complex hierarchical systems
of interpretation, the genetic information can move from a
more-or-less direct representation of physical constraints to a
representation in progressively more abstract and sophisticated
languages of interpretation: cf.\
\citep{pattee1972laws,pattee2015cell},
\citep{barricelli1987suggestions}.  
In the context of procedural content generation for computer games,
Michael Cook discusses the distinction between \emph{possibility space}
(defined as the set of all possible worlds we can image, represent or
describe) and \emph{generative space} (the set of all things a given
generative system can produce), where the latter is a subset of the
former.\footnote{\url{https://www.possibilityspace.org/tutorial-generative-possibility-space/}}
We can usefully borrow these concepts to say that the evolution of
higher-level, more complex systems of interpretation of the genetic
information into physical constraints and dynamics results in the
creation of new generative spaces that increase the likelihood of
generating complex, adaptive organisms and higher-level persistent
systems.

\section{Looking Forward}
In the sections above I have suggested that evolutionary innovations
can be understood in terms of capturing novel physical phenomena and of
system building. Of course, I do not claim that this is necessarily
the \emph{best} way of viewing innovations---that will depend upon one's
goals. Nevertheless, it is an alternative perspective that connects
questions of evolutionary innovation to existing literature on
systemic approaches to understanding biological and technological
phenomena. 

Viewing evolution as the process of selecting persistent
information-driven systems offers the potential of a consistent
description of evolutionary innovations at all levels, from cells to
ecosystems and cultures. The successful development of this approach
will depend upon the elaboration of the kind of general systems
vocabulary proposed by \cite{rosnay1975macroscope} and others into a
more formal and comprehensive general descriptive language of system
design and function.

The systems view gives a general description of the types of
novelties described in \citep{taylor2019evolutionary} as
\emph{non-additive compositional systems} and suggests three distinct
ways in which these kinds of novelties can arise, as listed on the previous page.
These kinds of novelty, together with the discovery of novel physical
phenomena, allow evolutionary systems to expand their search spaces
and thereby have the potential for interesting (i.e.\ \emph{expansive}
and \emph{transformational}) types of open-endedness rather than mere
\emph{exploratory} open-endedness in a fixed search space.

If successfully developed, a general systems view on evolutionary
innovations would be useful in categorising novelties across many
different types of open-ended system and in suggesting mechanisms for
designing new artificial worlds with improved open-ended evolutionary
potential.

\footnotesize
\bibliographystyle{apalike}
\bibliography{taylor-oee4} 

\begin{thebibliography}{}

\bibitem[Banzhaf et~al., 2016]{banzhaf2016defining}
Banzhaf, W., Baumgaertner, B., Beslon, G., Doursat, R., Foster, J.~A.,
  McMullin, B., De~Melo, V.~V., Miconi, T., Spector, L., Stepney, S., et~al.
  (2016).
\newblock Defining and simulating open-ended novelty: requirements, guidelines,
  and challenges.
\newblock {\em Theory in Biosciences}, 135(3):131--161.

\bibitem[Barricelli, 1987]{barricelli1987suggestions}
Barricelli, N.~A. (1987).
\newblock Suggestions for the starting of numeric evolution processes to evolve
  symbioorganisms capable of developing a language and technology of their own.
\newblock {\em Theoretic Papers}, 6(6):119--146.
\newblock (A publication of the Blindern Theoretic Research Team, University of
  Oslo).

\bibitem[Baugh and McMullin, 2013]{baugh2013evolution}
Baugh, D. and McMullin, B. (2013).
\newblock Evolution of {G-P} mapping in a von {N}eumann self-reproducer within
  {T}ierra.
\newblock In {\em ECAL 2013: The Twelfth European Conference on Artificial
  Life}, pages 210--217. MIT Press.

\bibitem[Bechtel, 2007]{bechtel2007biological}
Bechtel, W. (2007).
\newblock Biological mechanisms: Organized to maintain autonomy.
\newblock In Boogerd, F.~C., Bruggeman, F.~J., Hofmeyr, J.-H.~S., and
  Westerhoff, H.~V., editors, {\em Systems Biology}, pages 269--302. Elsevier.

\bibitem[Bird and Layzell, 2002]{bird2002evolved}
Bird, J. and Layzell, P. (2002).
\newblock The evolved radio and its implications for modelling the evolution of
  novel sensors.
\newblock In {\em Proceedings of the 2002 Congress on Evolutionary Computation.
  CEC'02 (Cat. No. 02TH8600)}, volume~2, pages 1836--1841. IEEE.

\bibitem[Cariani, 1993]{cariani1993evolve}
Cariani, P. (1993).
\newblock To evolve an ear. epistemological implications of {G}ordon {P}ask's
  electrochemical devices.
\newblock {\em Systems Research}, 10(3):19--33.

\bibitem[{de Rosnay}, 1975]{rosnay1975macroscope}
{de Rosnay}, J. (1975).
\newblock {\em Le macroscope, vers une vision globale}.
\newblock {\'E}ditions du Seuil, Paris.
\newblock English version entitled ``The macroscope : a new world scientific
  system'' first published in 1975 by Harper \& Row, New York.

\bibitem[Grand, 2000]{grand2000creation}
Grand, S. (2000).
\newblock {\em Creation: Life and how to make it}.
\newblock Weidenfeld \& Nicolson, London.

\bibitem[Kauffman, 2020]{kauffman2020answering}
Kauffman, S. (2020).
\newblock Answering {S}chr{\"o}dinger’s "{W}hat {I}s {L}ife?".
\newblock {\em Entropy}, 22(8):815.

\bibitem[Lenton et~al., 2021]{lenton2021survival}
Lenton, T.~M., Kohler, T.~A., Marquet, P.~A., Boyle, R.~A., Crucifix, M.,
  Wilkinson, D.~M., and Scheffer, M. (2021).
\newblock Survival of the systems.
\newblock {\em Trends in Ecology \& Evolution}, 36(4):333--344.

\bibitem[Marletto, 2015]{marletto2015constructor}
Marletto, C. (2015).
\newblock Constructor theory of life.
\newblock {\em Journal of The Royal Society Interface}, 12(104):20141226.

\bibitem[Mont{\'e}vil and Mossio, 2015]{montevil2015biological}
Mont{\'e}vil, M. and Mossio, M. (2015).
\newblock Biological organisation as closure of constraints.
\newblock {\em Journal of Theoretical Biology}, 372:179--191.

\bibitem[Pahl and Beitz, 1988]{pahl1988engineering}
Pahl, G. and Beitz, W. (1988).
\newblock {\em Engineering design : a systematic approach}.
\newblock Design Council, London.

\bibitem[Pattee, 2015]{pattee2015cell}
Pattee, H. (2015).
\newblock Cell phenomenology: The first phenomenon.
\newblock {\em Progress in Biophysics and Molecular Biology}, 119(3):461--468.

\bibitem[Pattee, 1972]{pattee1972laws}
Pattee, H.~H. (1972).
\newblock Laws and constraints, symbols and languages.
\newblock In Waddington, C.~H., editor, {\em Towards a Theoretical Biology},
  volume~4, pages 248--258. Edinburgh University Press.

\bibitem[Ray, 1991]{ray1991approach}
Ray, T.~S. (1991).
\newblock An approach to the synthesis of life.
\newblock {\em Artificial life II}, pages 371--408.

\bibitem[Sloman, 2017]{sloman2017construction}
Sloman, A. (2017).
\newblock Construction kits for biological evolution.
\newblock In Cooper, S.~B. and Soskova, M.~I., editors, {\em The Incomputable:
  Theory and Applications}, pages 237--292. Springer.

\bibitem[Sterman, 2000]{sterman2000business}
Sterman, J.~D. (2000).
\newblock {\em Business Dynamics: Systems Thinking and Modeling for a Complex
  World}.
\newblock McGraw-Hill.

\bibitem[Taylor, 2019]{taylor2019evolutionary}
Taylor, T. (2019).
\newblock Evolutionary innovations and where to find them: Routes to open-ended
  evolution in natural and artificial systems.
\newblock {\em Artificial Life}, 25(2):207--224.

\bibitem[{von Bertalanffy}, 1968]{bertalanffy1968general}
{von Bertalanffy}, L. (1968).
\newblock {\em General system theory: foundations, development, applications}.
\newblock George Braziller, New York.

\bibitem[Wiener, 1948]{wiener1948cybernetics}
Wiener, N. (1948).
\newblock {\em Cybernetics: or Control and Communication in the Animal and the
  Machine}.
\newblock Wiley, New York.

\end{thebibliography}

\end{document}